%% file: main.tex
\pdfoutput=1
\documentclass[11pt]{article}

\PassOptionsToPackage{table}{xcolor}

\usepackage[final]{acl}

\usepackage{times}
\usepackage{latexsym}
\usepackage[T1]{fontenc}
\usepackage[utf8]{inputenc}
\usepackage{microtype}
\usepackage{inconsolata}
\usepackage{url}

\usepackage{amsmath,amssymb,mathtools,amsthm}
\usepackage{algorithm}
\usepackage{algpseudocode}

\usepackage{graphicx}
\usepackage{subcaption}
\usepackage{booktabs}
\usepackage{multirow}
\usepackage{array}
\usepackage{tabularx}
\usepackage{longtable}
\usepackage{adjustbox}
\usepackage{wrapfig}
\usepackage{tablefootnote}
\usepackage{bbm}
\usepackage{float}
\usepackage{enumitem}
\usepackage[most]{tcolorbox}

\usepackage{titletoc}

\usepackage[capitalize,noabbrev]{cleveref}

\input{header}

\title{Prune the Summary, Not the History: \\Source-Judged Compaction for Long-Horizon Agents}

\author{Guanghui Min \\
  Department of Computer Science \\
  University of Virginia \\
  \texttt{jjm8vr@virginia.edu}
  \And
  Chen Chen\thanks{Corresponding author.} \\
  Department of Computer Science \\
  University of Virginia \\
  \texttt{zrh6du@virginia.edu}}

\begin{document}
\maketitle
\begin{abstract}
\input{sections/0-abstract}
\end{abstract}

\section{Introduction}

\input{sections/1-introduction}

\section{Preliminary}
\input{sections/2-prelimiary}

\section{Why Summary Needs Source Judgment}
\label{sec:motivation}

\input{sections/3-motivation}

\section{\texttt{SJC}: Our Proposed Method}

\input{sections/4-methodology}

\section{Experiments}\label{sec:experiments}
\input{sections/5-experiment}

\section{Related Work}
\input{sections/related_work}

\section{Conclusion}
\input{sections/6-conclusion}

\newpage
\section*{Limitations}

\texttt{SJC} filters by provenance, not correctness, and the two are
largely independent: supported and unsupported entries are correct at
similar rates ($92.2\%$ against $90.6\%$;
Appendix~\ref{app:judge}). An entry that cites a real observation but
draws a wrong conclusion from it has a source and is kept: in our
analysis of failed runs, four of the $29$ entries that the agent acted
on were supported by environment outputs, and no source-based filter
would remove them. Conversely, \texttt{SJC} removes entries that are
correct and sometimes entries that help. At our threshold the judge
errs toward pruning and recovers $64.2\%$ of supported entries. On one
AppWorld task it prunes a compressor-written note restricting a text
message to newly registered recipients; GPT-5.6 Luna recovers the
restriction from the task instruction, but MiniMax-M3 does not and
fails in two of three runs.

Our evaluation covers two benchmarks, two agent models, and one host
compressor, with compaction triggered at small context windows so that
it occurs within each task. In both benchmarks the task is a single
instruction that stays visible after compaction, which is what makes
it safe to remove entries that only restate it. In multi-turn settings,
earlier user messages are themselves a source and would need to be
treated as one; we have not evaluated that case. \texttt{SJC} also
depends on an external judge model accessed through an API, whose
scores are not fully deterministic, and replacing it with the
compressor itself was less reliable in our experiments. Finally, the
failure analysis in Section~\ref{sec:motivation} relies on annotations
produced by LLM annotators under a written protocol.




\bibliography{main}

\appendix

\newpage

\input{sections/appendix}


\end{document}

%% file: header.tex
\newcommand{\yes}{\textcolor{green!60!black}{\ding{51}}}
\newcommand{\no}{\textcolor{red!75!black}{\ding{55}}}

\usepackage{pifont}

\definecolor{ICMLBlue}{rgb}{0.0, 0.0, 0.5}
\definecolor{CiteTeal}{HTML}{008080}
\definecolor{lightcrimson}{rgb}{0.93, 0.16, 0.51}
\definecolor{LinkRed}{rgb}{0.768, 0.054, 0.054}

\definecolor{lightgray}{gray}{0.9}
\definecolor{lightblue}{RGB}{230,245,255}

\theoremstyle{plain}

\theoremstyle{definition}

\theoremstyle{remark}

\titlecontents{section}[1.5em]
    {\addvspace{6pt}\bfseries}
    {\contentslabel{1.5em}}
    {\hspace*{-1.5em}}
    {\titlerule*[0.5pc]{.}\contentspage}

\titlecontents{subsection}[3.8em]
    {}
    {\contentslabel{2.3em}}
    {\hspace*{-2.3em}}
    {\titlerule*[0.5pc]{.}\contentspage}

\newcommand{\std}[1]{\textcolor{gray}{\fontsize{5}{6}\selectfont$\pm$#1}}

\makeatletter
\@ifpackageloaded{algorithm}{
    
}{}
\makeatother

\tcbuselibrary{breakable,skins,listings}

\definecolor{BoxGrayBack}{HTML}{F8F8F8}
\definecolor{BoxGrayFrame}{HTML}{B8B8B8}
\definecolor{BoxBlueBack}{HTML}{F5F9FF}
\definecolor{BoxBlueFrame}{HTML}{2D5DA8}
\definecolor{BoxGreenBack}{HTML}{F6FFF8}
\definecolor{BoxGreenFrame}{HTML}{2E7D32}

\newtcolorbox{detailbox}[2][]{
  enhanced,
  breakable,
  colback=BoxGrayBack,
  colframe=BoxGrayFrame,
  boxrule=0.5pt,
  arc=2pt,
  left=6pt,
  right=6pt,
  top=5pt,
  bottom=5pt,
  fonttitle=\bfseries,
  coltitle=black,
  title={#2},
  #1
}

\newtcblisting{promptbox}[2][]{
  enhanced,
  breakable,
  listing only,
  colback=BoxBlueBack,
  colframe=BoxBlueFrame,
  boxrule=0.6pt,
  arc=2pt,
  left=6pt,
  right=6pt,
  top=5pt,
  bottom=5pt,
  fonttitle=\bfseries,
  coltitle=white,
  colbacktitle=BoxBlueFrame,
  title={#2},
  listing options={
    basicstyle=\ttfamily\footnotesize,
    breaklines=true,
    breakatwhitespace=false,
    columns=fullflexible,
    keepspaces=true,
    showstringspaces=false,
    upquote=true
  },
  #1
}

\newtcolorbox{implementationbox}[2][]{
  enhanced,
  breakable,
  colback=BoxGreenBack,
  colframe=BoxGreenFrame,
  boxrule=0.6pt,
  arc=2pt,
  left=6pt,
  right=6pt,
  top=5pt,
  bottom=5pt,
  fonttitle=\bfseries,
  coltitle=white,
  colbacktitle=BoxGreenFrame,
  title={#2},
  #1
}

\newtcolorbox{takeawaybox}{
  colback=red!4,
  colframe=red!60!black,
  boxrule=0.65pt,
  arc=1.5pt,
  left=6pt,
  right=6pt,
  top=5pt,
  bottom=5pt
}


%% file: sections/0-abstract.tex
Long-horizon LLM agents rely on context compaction: when the
interaction history outgrows the context window, a compressor replaces
it with a natural-language summary. We show that these summaries
contain more than a record of what happened. Compaction prompts ask
for decisions and next steps, so each summary mixes facts established
by the environment with reasoning written by the compressor, and the
agent cannot tell the two apart afterwards. In an analysis of a widely
used structured compressor, $83\%$ of failed runs trace back to an
action that carries out a specific summary entry, most of which no
environment output supports. We propose Source-Judged Compaction
(\texttt{SJC}), a training-free post-processing step that scores each
summary entry for whether it has a source in the environment outputs
or the previous summary, and removes the entries that do not. It
changes neither the agent nor the compressor and adds one inexpensive
judge request per compaction. The judge is Jev, a non-generative
decision model that answers the source question with a calibrated
probability instead of text. On AppWorld and OfficeBench with two
agent models, \texttt{SJC} improves cross-run reliability
over the underlying compressor by $2.1\%$--$8.9\%$ without increasing
the agent's token usage. Our code is available at \url{https://github.com/GuanghuiMin/sjc}.

%% file: sections/1-introduction.tex
Long-horizon agentic tasks require LLMs to plan, use tools, and adapt to environmental feedback across extended interactions~\citep{xu2024theagentcompany,wang2024officebench}. During execution, tool outputs, failed attempts, intermediate results, and reasoning traces continually accumulate, making the context increasingly expensive to process and difficult to navigate. This problem persists even with large nominal context windows, whose effective use of distant information degrades with context length~\citep{liu2024lostmiddle,hsieh2024ruler}. Context compaction is therefore essential for scalable long-horizon agents.

General compaction methods select or rewrite tokens from static prompts and documents~\citep{jiang2023llmlingua,pan2024llmlingua}, but do not directly account for subsequent agent execution. Agent-specific methods such as ReSum, SUPO, and ACM instead modify or post-train the agent to manage compressed contexts~\citep{wu2025resum,lu2025supo,li2026acm}. In many deployed systems, however, the agent is fixed or accessible only through an API. OpenClaw, DeepSeek Harness, and Codex therefore invoke a separate compressor that replaces older interactions with a natural-language checkpoint~\citep{openclaw_compaction,deepseek_harness_compaction,openai_codex_compaction}, while ACON and PAIR improve the compression prompt without modifying the downstream agent~\citep{kang2025acon,min2026adapting}.

Such checkpoints do more than shorten history: they periodically
reconstruct the state from which the agent continues, including its
progress, constraints, and future actions. Errors in this
reconstruction can therefore alter subsequent
execution~\citep{min2026toward,min2026adapting}. For example, when a
user asks to reply to a friend with movie titles ``as per their
request'', the compressor records the next step as sending the
comma-separated movie list; the agent then sends every title it
found, including those the friend did not ask for. The compressor
wrote this plan itself, but after compaction the agent cannot tell it
from a fact the environment established. Such cases are common: in
our analysis of a widely used structured compressor, $67.7\%$ of the
pending work it records has no source in the environment outputs, and
$29$ of $35$ failed runs trace back to an action that carries out a
specific summary entry. This suggests filtering summary entries by
their provenance. Whether an entry is supported by an environment
output or by the previous summary can be checked against the history
at compaction time, without knowing how the rest of the task will
unfold.

We introduce \texttt{SJC} (\textbf{S}ource-\textbf{J}udged
\textbf{C}ompaction), a lightweight plug-in that prunes unsupported
entries from generated summaries. Given a structured natural-language
summary, \texttt{SJC} splits it into entries, retrieves relevant
environment outputs, and asks a probabilistic judge whether each entry
contains information that came from those outputs or from the previous
summary, as opposed to only plans, decisions, or interpretations of
the task. Entries without such a source are removed as complete units,
while retained entries remain identical to the compressor's output.
The pruned summary is then shown to the agent and used as the previous
summary at the next compaction. \texttt{SJC} requires no modification
to the agent, the compressor prompt, or the runtime, and adds no
summary-generation call. Our contributions are threefold:

\begin{itemize}[leftmargin=0.25cm, nosep]
    \item \textbf{Analysis of Compaction Summaries:} We survey 29
    compaction prompts from widely used agent systems and find that
    most ask the compressor for decisions and next steps. Analyzing
    the resulting summaries, we show that much of this reasoning has
    no source in the environment outputs, that it is regenerated at
    every compression boundary, and that most failed runs follow a
    specific summary entry.

    \item \textbf{Source-Judged Compaction:} We develop a
    training-free post-processing step that decomposes a structured
    summary into entries, retrieves relevant environment evidence, and
    removes the entries that a source judge finds unsupported. The
    method preserves the natural-language summary interface and
    attaches to existing agent systems without modifying them.

    \item \textbf{Empirical Effectiveness:} Across
    AppWorld~\citep{trivedi2024appworld} and
    OfficeBench~\citep{wang2024officebench}, using both GPT-5.6
    Luna~\citep{openai2026gpt56luna} and
    MiniMax-M3~\citep{minimax2026m3}, \texttt{SJC} improves cross-run
    reliability over the underlying prompting-based compressor in all
    four settings, with the largest gains on hard and
    multi-application tasks, without increasing the agent's token
    usage.
\end{itemize}

%% file: sections/2-prelimiary.tex
\paragraph{Notations.}
We consider an LLM agent $\pi$ that solves a task by interacting with
an environment over multiple steps. The agent is given a fixed prompt
$P$, which contains the system prompt and the task instruction and is
never compressed. At step $i$, the agent takes an action $a_i$ and
receives an environment output $o_i$; only $o_i$ is produced by the
environment, whereas $a_i$ and any accompanying reasoning are produced
by the agent. A rollout is $\rho=(a_1,o_1,\ldots,a_T,o_T)$, and
$S(\rho)\in\{0,1\}$ indicates whether it solves the task. We write
$|x|$ for the token length of $x$ and $\oplus$ for concatenation.

Because the interaction history grows with every step, agent systems
compact it whenever the context exceeds a budget $B$. We index these
compression boundaries by $t=1,2,\ldots$. At boundary $t$, the context
consists of the fixed prompt $P$, the previous summary $M_{t-1}$
(with $M_0=\varnothing$), the history span $H_t$ accumulated since the
last boundary, and the most recent interaction $R_t$, which is
retained verbatim. A compressor $\mathcal C$ replaces $M_{t-1}$ and
$H_t$ with a new summary $M_t$, and the agent continues from
$P\oplus M_t\oplus R_t$. We focus on structured natural-language
summaries, in which $M_t$ is a list of short entries grouped under
section headings.

\input{tables/intro}

\paragraph{Problem Definition.}
Compaction is recurrent: after boundary $t$, the summary $M_t$ is the
only record of $H_t$ available to the agent, and it is also the input
from which $M_{t+1}$ is produced. Any content written into a summary
is therefore treated as established by all later steps, whether or not
the environment ever produced it.
We study how to improve an existing compressor without changing it.
Let $\widetilde M_t=\mathcal C(P,M_{t-1},H_t,R_t)$ be the candidate
summary of a host compressor and $\mathcal B(\cdot)$ the set of
entries in a summary. We seek a post-processing operator $G$ that
commits $M_t=G(\widetilde M_t\,;P,M_{t-1},H_t,R_t)$ in place of
$\widetilde M_t$ so as to maximize task success,
\begin{equation}
\begin{aligned}
\max_{G}\quad
& \mathbb E_{\rho\sim p_{\pi,\mathcal C,G}}
\left[S(\rho)\right] \\
\text{s.t.}\quad
& \mathcal B(M_t)\subseteq\mathcal B(\widetilde M_t)
\quad \text{for all } t,
\end{aligned}
\label{eq:problem}
\end{equation}
where $p_{\pi,\mathcal C,G}$ is the rollout distribution induced by
the agent, the environment, the host compressor, and $G$. The
constraint restricts $G$ to deleting entries: it may not rewrite or
add content, so $|M_t|\leq|\widetilde M_t|$ and every retained entry
is exactly what the host compressor wrote. We further require $G$ to
leave $\pi$ and $\mathcal C$ unchanged and to use no training or
task-specific adaptation, so that it can be attached to any
compressor that produces structured summaries.

%% file: tables/intro.tex
\refstepcounter{footnote}
\edef\resumfootnote{\number\value{footnote}}

\begin{table}[t]
\centering
\footnotesize
\setlength{\tabcolsep}{1.5pt}
\renewcommand{\arraystretch}{1.06}

\begin{tabularx}{\linewidth}{
    @{}
    >{\raggedright\arraybackslash}X
    >{\centering\arraybackslash}p{0.16\linewidth}
    >{\centering\arraybackslash}p{0.21\linewidth}
    >{\centering\arraybackslash}p{0.16\linewidth}
    @{}
}
\toprule
\textbf{Method}
& \textbf{Summary}
& \textbf{Train-Free}
& \textbf{Drop-in} \\
\midrule

ACE~\citep{zhang2026agentic}
& \no & \yes & \no \\

Mem1~\citep{mem1}
& \no & \no & \no \\

ACM~\citep{li2026acm}
& \no & \no & \no \\

Fast-Jev~\citep{tran2026fastjev}
& \no & \yes & \yes \\

\midrule

ReSum\textsuperscript{\resumfootnote}~\citep{wu2025resum}
& \yes & \no & \no \\

SUPO~\citep{lu2025supo}
& \yes & \no & \no \\

ACON~\citep{kang2025acon}
& \yes & \yes & \no \\


\midrule

\textbf{SJC (Ours)}
& \yes & \yes & \yes \\

\bottomrule
\end{tabularx}

\caption{
Comparison of context-management approaches.
\textbf{Summary} denotes a natural-language history checkpoint;
\textbf{Train-Free} requires no context-management policy training;
and \textbf{Drop-in} requires neither agent modification nor
task-specific adaptation.
}
\label{tab:method_positioning}
\vspace{-5mm}
\end{table}

\footnotetext[\resumfootnote]{We consider the GRPO-trained ReSum variant.}

%% file: sections/3-motivation.tex
We analyze the summaries written by a widely used structured
compressor, OpenClaw~\citep{openclaw_compaction}, on the training
split of AppWorld~\citep{trivedi2024appworld} (90 tasks, three runs,
GPT-5.6 Luna). We call a summary entry \emph{supported} if its content
is established by an environment output or by the previous summary,
and \emph{unsupported} otherwise. To label entries at scale, we ask
Jev~\citep{typesafe2026jev} whether each entry contains information
from either source and treat entries scoring below $0.5$ as
unsupported (Appendix~\ref{app:composition}). Against annotated labels, Jev separates the two with an AUROC of
$0.905$; it is conservative at this threshold, so the unsupported
shares we report are upper bounds (Appendix~\ref{app:judge}).

\subsection{Summary as Periodic Reasoning}
\label{sec:periodic-reasoning}

A compaction prompt asks for more than a record of what happened. We
collected 29 compaction prompts from the official repositories of
widely used agent systems (Appendix~\ref{app:prompts}). Of these, 21
ask the compressor for decisions or next steps: 18 request the
decisions made and 17 request the remaining work, in instructions such
as ``What remains to be done (clear next steps)'' in Codex CLI and
``The current plan and the IMMEDIATE next step'' in Gemini CLI.
OpenClaw requires sections for \textit{Decisions}, \textit{Open
TODOs}, and \textit{Pending user asks}. Each compression boundary is
therefore also a reasoning step: the compressor re-derives where the
task stands and where it should go next. If environment outputs
describe the state of the task, this reasoning describes its
direction, which the environment never reported.

\noindent\textbf{Much of this reasoning is unsupported.}
Table~\ref{tab:composition} breaks down the $26{,}881$ entries of
$667$ summaries by section. Sections that record observations are
almost entirely supported, whereas $67.7\%$ of open TODOs and $86.1\%$
of pending user asks have no source. Among the entries under
\textit{Decisions} and \textit{Open TODOs}, which we call reasoning
entries, $38.7\%$ are unsupported, and $54.9\%$ at the first
compression boundary.

\noindent\textbf{Unsupported reasoning is regenerated at every
boundary.}
These entries are not simply carried from one summary to the next:
only $20.5\%$ of unsupported reasoning entries reappear in the
following summary, and $5.3\%$ three summaries later. Yet every
summary contains about four of them, whether it is the first
($4.3$), the second ($3.9$), or a later one ($3.8$). The compressor
writes unsupported reasoning anew at each boundary, so removing it
once is not enough.

\begin{table}[t]
\centering
\small
\setlength{\tabcolsep}{5pt}
\caption{Entries of OpenClaw summaries on the AppWorld training split,
by section. An entry is unsupported when Jev finds no environment
output or previous summary that establishes it.}
\label{tab:composition}
\begin{tabular}{lrr}
\toprule
Section & Entries & Unsupported (\%) \\
\midrule
Exact identifiers & 14{,}444 & 3.4 \\
Constraints/Rules & 4{,}183 & 11.1 \\
Decisions & 4{,}270 & 21.0 \\
Open TODOs & 2{,}600 & 67.7 \\
Pending user asks & 1{,}384 & 86.1 \\
\midrule
All & 26{,}881 & 17.9 \\
\bottomrule
\end{tabular}
\vspace{-3mm}
\end{table}

\begin{figure*}[t]
    \centering
    \includegraphics[width=\linewidth]{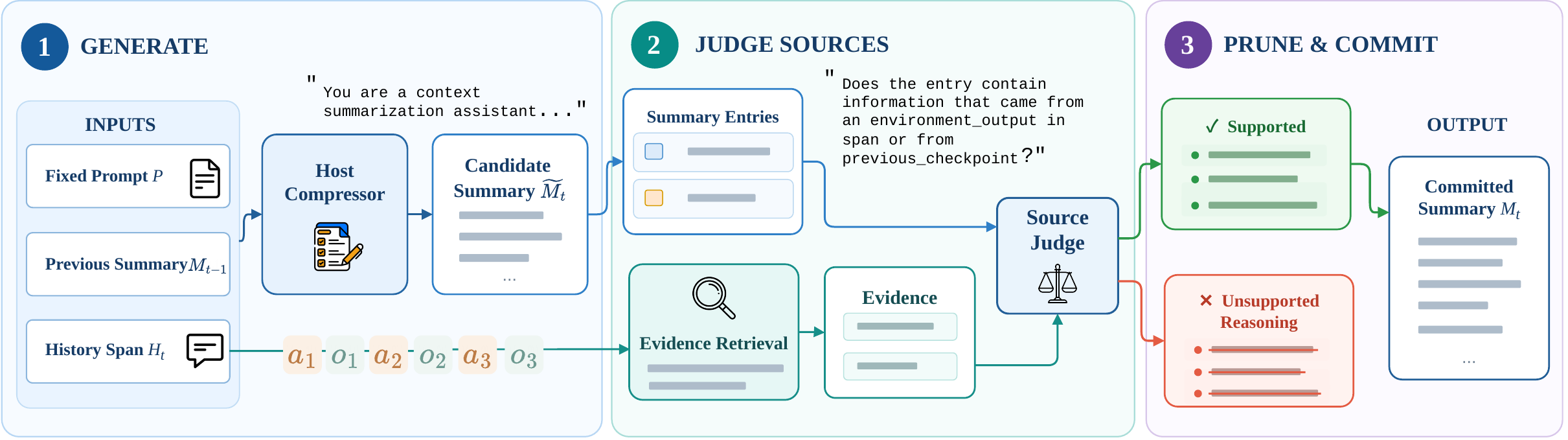}
    \caption{
        \textbf{Overview of \texttt{SJC}.}
        (1) A host compressor produces a structured candidate summary.
        (2) \texttt{SJC} splits it into entries, retrieves relevant
        environment evidence from the history span, and scores all
        entries with a source judge in a single request.
        (3) Supported entries are retained unchanged, while
        unsupported entries are removed before the committed summary
        is shown to the agent and passed to the next compression
        boundary.
    }
    \label{fig:sjc-framework}
    \vspace{-4mm}
\end{figure*}

\subsection{Unsupported Reasoning Can Derail Agent Execution}
\label{sec:derail}

After compaction, the agent cannot tell which entries the environment
established and which the compressor inferred; it treats both as
facts. Developers have met this problem in deployment: Hermes prepends
to every summary a notice telling the agent to ``treat it as
background reference, NOT as active
instructions''~\citep{hermes_context_compression}. To see what the
problem costs, we trace every failed run in our analysis back from the
failed evaluation check to the action that caused it
(Appendix~\ref{app:failures}). Of the $44$ failed runs, nine belong to
a scenario whose reference solution is inconsistent with its
environment data and that no method solves; we analyze the remaining
$35$.

\noindent\textbf{Most failures follow a summary entry.}
In $29$ of the $35$ failed runs, the failing action carries out a
specific summary entry; two independent rounds of annotation agree on
every run. Read against the environment outputs that preceded them,
$22$ of these entries are supported by no output and $3$ contradict
one; only $4$ are supported. The failed runs span $10$ task scenarios;
counted once per scenario and failure pattern, $15$ of $19$ cases
follow an entry.

\noindent\textbf{The failures fall into a few recurring patterns.}
\textit{Rewritten constraints} ($15$ runs) restate the task with a
different meaning. In one task, the user asks to reply to a friend
with movie titles ``as per their request''; the summary records the
TODO ``Send Leslie the comma-separated movie list by text message'',
and the agent sends every title in the note, including those the
friend did not ask for.
\textit{Invented deliverables} ($5$) tell the agent to report a
result, which it then submits as an answer the task does not expect.
\textit{Contradicted state} ($3$) asserts something an environment
output refutes. The remaining $6$ carry forward an earlier choice of
the agent or present a partial result as complete.

Both observations point to the same remedy: check each summary entry
against its source, and do so at every compression boundary.

%% file: sections/4-methodology.tex
Figure~\ref{fig:sjc-framework} illustrates \texttt{SJC}
(\textbf{Source-Judged Compaction}), a training-free post-processing
method for structured natural-language compaction. A host compressor
first produces a summary in its original format. \texttt{SJC} then
judges the source of each summary entry and removes entries without
an identifiable source. It neither rewrites retained entries nor
changes the agent or the host compressor.

\noindent\textbf{Setup.}
At compression boundary $t$, let $P$ denote the fixed prompt retained
verbatim, $H_t$ the history span being compressed, $R_t$ the most
recent interaction retained verbatim, and $M_{t-1}$ the previous
committed summary. \texttt{SJC} proceeds in three stages and outputs a
committed summary $M_t$, which serves as the agent's checkpoint: it is
shown to the agent and passed to the compressor at the next boundary.

\noindent\textbf{Stage 1: Generate.}
The host compressor $\mathcal C$ summarizes the history span as a
structured natural-language candidate summary:
\begin{equation}
\widetilde M_t
=
\mathcal C(P,M_{t-1},H_t,R_t).
\label{eq:candidate-summary}
\end{equation}
Unlike methods that directly select raw history chunks, \texttt{SJC}
retains this summary interface because it provides the agent with a
compact and readable description of the current task state.

\noindent\textbf{Stage 2: Judge sources.}
\texttt{SJC} parses the candidate summary into entries
$\mathcal B_t=\{b_{t,1},\ldots,b_{t,N_t}\}$ while preserving their
section headings. For each entry, it retrieves a small set of relevant
passages from the environment outputs in $H_t$ and pools them into a
shared evidence set:
\begin{equation}
\mathcal E_t
=
\bigcup_{k=1}^{N_t}
\operatorname{Retrieve}(b_{t,k},H_t).
\label{eq:evidence-retrieval}
\end{equation}
The pooled evidence is much shorter than $H_t$, so the judge never
needs to read the full history. Retrieval affects only the evidence
shown to the judge; it does not directly determine which entries are
kept.
A source judge $J$ then receives all entries together with the pooled
evidence in a single request and returns one score per entry:
\begin{equation}
\left(p_{t,1},\ldots,p_{t,N_t}\right)
=
J\!\left(
\mathcal B_t
\mid
P,R_t,M_{t-1},\mathcal E_t
\right),
\label{eq:source-judgment}
\end{equation}
where $p_{t,k}$ is the judge's confidence that entry $b_{t,k}$
contains information that came from an environment output or from the
previous committed summary. \texttt{SJC} thus adds one judge request
per boundary, however many entries there are.

The judgment concerns provenance rather than general reasoning
quality. Values, names, observed results, and established environment
behavior can be traced to environment outputs or the previous
summary. In contrast, plans, intentions, and interpretations produced
only by the agent or the compressor lack such support. The fixed
prompt $P$ is given to the judge as context but does not count as a
source: it remains visible to the agent after compaction, so entries
that only restate or interpret the task need not be committed to the
summary.

\noindent\textbf{Stage 3: Prune and commit.}
\texttt{SJC} retains an entry when its source score reaches a fixed
threshold $\tau$:
\begin{equation}
g_{t,k}
=
\mathbbm{1}\!\left[p_{t,k}\geq\tau\right].
\label{eq:source-gate}
\end{equation}
The committed summary is
\begin{equation}
M_t
=
\operatorname{Render}
\left(
\left\{
b_{t,k}
\in\mathcal B_t
:
g_{t,k}=1
\right\}
\right).
\label{eq:committed-summary}
\end{equation}
Unsupported entries are removed as complete units. Supported entries
remain identical to the host compressor's output, and the summary
structure is preserved.
After compaction, the agent receives
\begin{equation}
X_t = P \oplus M_t \oplus R_t.
\label{eq:agent-context}
\end{equation}
The same $M_t$ is used as the previous summary at the next compression
boundary. Because the compressor writes unsupported entries anew at
each boundary (Section~\ref{sec:periodic-reasoning}), \texttt{SJC}
applies this gate at every boundary rather than once. It preserves the
natural-language summary produced by the host compressor while keeping
unsupported content out of the context the agent sees.

%% file: sections/5-experiment.tex
\input{tables/gpt_minimax_appworld}

\input{tables/gpt_minimax_officebench}

\subsection{Experimental Setup}

\paragraph{Datasets.}
We evaluate on two long-horizon agent benchmarks.
\textbf{AppWorld}~\citep{trivedi2024appworld} contains stateful
API-use tasks across simulated applications and users; we use its
168-task test-normal split.
\textbf{OfficeBench} covers multi-application office workflows
involving documents, spreadsheets, and email
\citep{wang2024officebench}; we use the 95-task test set of
ACON~\citep{kang2025acon}.

\paragraph{Metrics.}
Following \citet{min2026adapting}, we evaluate each method over three independent runs. \textbf{Acc.} denotes the mean task success rate, while \textbf{Pass$^3$} denotes the fraction of tasks solved in all three runs and measures cross-run reliability. We also report the mean number of interaction \textbf{Steps} and the peak input length (\textbf{Peak}), measured in thousands of tokens. For efficiency, we report cumulative agent and compressor tokens, including all inputs and outputs, and report the API cost of Jev separately.

\paragraph{Models.}
We use GPT-5.6 Luna~\citep{openai2026gpt56luna} (with medium reasoning
effort) and MiniMax-M3~\citep{minimax2026m3} as both the agent and the
compressor. For each configuration, the agent and compressor use the
same model, following the standard design of current agent systems.
The source judge is Jev-1.13~\citep{typesafe2026jev}, a non-generative
decision model that answers a typed question with a calibrated
probability rather than free-form text. We pose the source question in
Figure~\ref{fig:sjc-framework} as a yes-or-no decision and use the returned
probability of ``yes'' as $p_{t,k}$ in Eq.~\eqref{eq:source-judgment}. We access
GPT-5.6 Luna and MiniMax-M3 through the official OpenAI and MiniMax
APIs, respectively, and Jev through the OpenRouter API.

\paragraph{Baselines.}
We use \textbf{No Compression}, which retains the full interaction
history, as the uncompressed reference. We compare against five
compression baselines: \textbf{(1) FIFO} retains the most recent
history within the context budget; \textbf{(2)
LLMLingua}~\citep{jiang2023llmlingua} and \textbf{(3)
LLMLingua-2}~\citep{pan2024llmlingua} perform token-level pruning using
learned importance models; \textbf{(4)
Fast-Jev}~\citep{tran2026fastjev} uses Jev to select raw tool calls
and outputs for retention, without producing a summary; and
\textbf{(5) Prompting} uses the original OpenClaw structured compaction
prompt~\citep{openclaw_compaction}. Our method, \textbf{Prompting +
SJC}, uses the same Jev model to prune unsupported entries from the
summaries produced by Prompting.

\paragraph{Compression Scope.}
Following OpenClaw~\citep{openclaw_compaction}, compaction retains a configurable number of messages from the beginning and recent interactions from the end of the context, while pruning or summarizing the intervening history. For simplicity, our experiments retain one item from each side: the first user message and the most recent agent action--observation interaction. Context windows, retrieval settings, and the judge configuration are
given in Appendix~\ref{app:implementation}.

\subsection{Main Results}

\texttt{SJC} achieves the highest Pass$^3$ among compressed methods in
all four benchmark--model settings, improving over Prompting by
$2.1\%$--$8.9\%$ without adding interaction steps or context length
(Tables~\ref{tab:appworld_gpt_luna}
and~\ref{tab:officebench_gpt_luna}). The gains concentrate on hard and
multi-application tasks, where compaction degrades execution the most.
Fast-Jev uses the same judge model to select raw history and falls
below Prompting in three of four settings, so the gain comes from what
the judge is asked and where it is applied, not from the judge itself.
All differences are reported in absolute terms.

\noindent\textbf{\texttt{SJC} improves cross-run reliability in every
setting.}
Pass$^3$ rises over Prompting by $8.9\%$ and $7.1\%$ on AppWorld
(GPT-5.6 Luna and MiniMax-M3) and by $5.3\%$ and $2.1\%$ on
OfficeBench. Accuracy rises by $1.1\%$--$5.3\%$ in three settings; on
OfficeBench with MiniMax-M3 it falls by $1.0\%$.

\noindent\textbf{The gains are largest where compaction hurts most.}
On AppWorld, Pass$^3$ improves by $10.4\%$ and $12.7\%$ on medium and
hard tasks with GPT-5.6 Luna, and by $10.4\%$ and $6.3\%$ with
MiniMax-M3. On OfficeBench 3-APP tasks with GPT-5.6 Luna, Pass$^3$
rises from $41.9\%$ to $54.8\%$ and accuracy from $59.1\%$ to
$64.5\%$. Across the twelve benchmark--model--subset groups, Pass$^3$
improves in ten, is unchanged in one, and decreases in one
(OfficeBench 1-APP with MiniMax-M3, $90.5\%$ to $88.1\%$), where
Prompting is near saturation.

\noindent\textbf{The gains require neither longer contexts nor
additional interaction.}
Relative to Prompting, \texttt{SJC} adds at most $0.1$ interaction
steps on average and changes peak input length by at most $0.47$K
tokens. Its peak input remains $27\%$--$54\%$ below uncompressed
execution. Section~\ref{sec:efficiency} reports the additional cost of
the judge.

\subsection{Efficiency}\label{sec:efficiency}

\begin{figure}[t]
\centering
\includegraphics[width=\columnwidth]{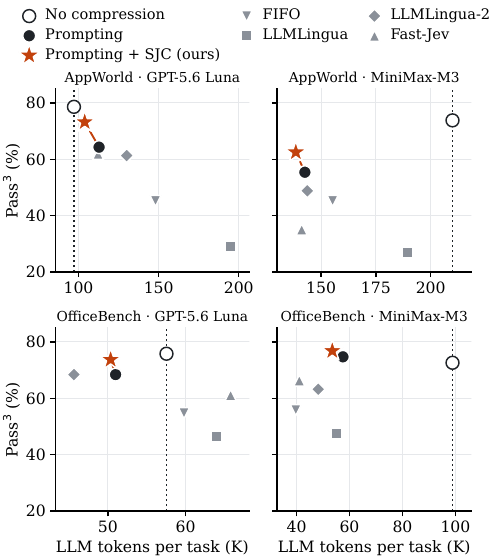}
\caption{Pass$^3$ versus mean cumulative LLM tokens per task (agent
plus LLM compressor). The dotted line marks the cost of uncompressed
execution. Jev calls are excluded and reported in
Table~\ref{tab:overhead}.}
\vspace{-5mm}
\label{fig:efficiency}
\end{figure}

\texttt{SJC} does not increase the cumulative LLM tokens of Prompting
in any setting, and its judge costs under \$0.002 per task
(Figure~\ref{fig:efficiency}, Table~\ref{tab:overhead}).

\noindent\textbf{\texttt{SJC} is no more expensive than the summaries
it prunes.}
Cumulative agent and compressor tokens change by $-8.0\%$ and $-2.9\%$
on AppWorld and by $-1.3\%$ and $-7.0\%$ on OfficeBench relative to
Prompting. The savings come from shorter summaries: pruning removes
$13\%$--$21\%$ of entries and shortens each summary by
$14\%$--$27\%$.

\noindent\textbf{The judge is cheap and its input is bounded.}
\texttt{SJC} issues one Jev request per compaction, covering all
entries of the candidate summary with their retrieved evidence.
No request exceeds $29.0$K input tokens, within Jev's $32$K context
window. At \$0.042 per million input tokens, the judge costs
\$1.2--\$1.5 per $1{,}000$ tasks.

\noindent\textbf{Compaction is cheaper than full history in three of
four settings.}
Relative to uncompressed execution, \texttt{SJC} uses $34\%$, $12\%$,
and $46\%$ fewer tokens on AppWorld with MiniMax-M3 and on OfficeBench
with both models. On AppWorld with GPT-5.6 Luna it uses $6.9\%$ more,
because compaction adds $5.8$ steps per task; \texttt{SJC} recovers
more than half of the $16.1\%$ overhead of Prompting there.

\begin{table}[t]
\centering
\small
\setlength{\tabcolsep}{4pt}
\caption{Overhead of \texttt{SJC}. LLM tokens are means per task;
$\Delta$ is relative to Prompting. Summary length is the change in
summary tokens after pruning.}
\label{tab:overhead}
\begin{tabular}{lcccc}
\toprule
& \multicolumn{2}{c}{AppWorld} & \multicolumn{2}{c}{OfficeBench} \\
\cmidrule(lr){2-3} \cmidrule(lr){4-5}
& GPT & M3 & GPT & M3 \\
\midrule
Prompting tokens (K) & 112.8 & 142.6 & 51.0 & 57.5 \\
\texttt{SJC} tokens (K) & 103.8 & 138.5 & 50.4 & 53.5 \\
$\Delta$ (\%) & $-8.0$ & $-2.9$ & $-1.3$ & $-7.0$ \\
\midrule
Entries pruned (\%) & 18.5 & 13.6 & 21.3 & 13.0 \\
Summary length (\%) & $-22.9$ & $-13.7$ & $-26.7$ & $-17.3$ \\
\midrule
Max judge input (K) & 22.5 & 25.9 & 28.4 & 29.0 \\
Judge cost (\$/1K tasks) & 1.24 & 1.39 & 1.19 & 1.52 \\
\bottomrule
\end{tabular}
\end{table}

\subsection{Sensitivity Analysis}

\begin{figure}[t]
\centering
\includegraphics[width=\columnwidth]{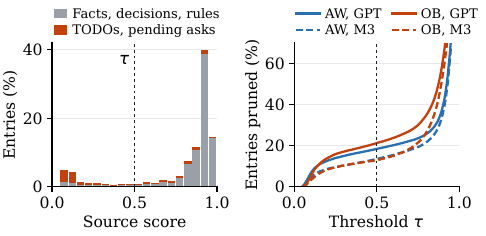}
\caption{Left: distribution of Jev source scores, with forward-looking
sections highlighted. Right: fraction of entries pruned as $\tau$
varies (AW: AppWorld; OB: OfficeBench). Dotted lines mark
$\tau = 0.5$.}
\vspace{-4mm}
\label{fig:sensitivity}
\end{figure}

\texttt{SJC} is insensitive to its threshold and prunes selectively
rather than by section; a binary self-judge is a less reliable
substitute for Jev.

\noindent\textbf{Pruning decisions are stable across thresholds.}
Jev scores are bimodal: $73.9\%$ of entries score at least $0.8$,
$10.6\%$ score below $0.2$, and only $8.1\%$ fall between $0.3$ and
$0.7$. Moving $\tau$ from $0.3$ to $0.7$ therefore changes the pruned
fraction by at most $10.2\%$ in any setting. We use $\tau = 0.5$
throughout.

\noindent\textbf{Pruning targets forward-looking entries without
removing them wholesale.}
\texttt{SJC} keeps $92.8\%$--$97.4\%$ of exact identifiers and
$84.2\%$--$92.5\%$ of decisions, but only $28.5\%$--$64.9\%$ of open
TODOs and $2.0\%$--$12.8\%$ of pending user asks. These two sections hold $15.2\%$ of all
entries and $59.5\%$ of pruned ones. Pruning is also concentrated at
the first compaction, which removes $29.9\%$--$39.6\%$ of entries
against $8.8\%$--$17.7\%$ at later ones, since entries carried over
from the previous checkpoint have already passed the judge.

\noindent\textbf{A calibrated non-generative judge is more repeatable than a generative self-judge.}
Replacing Jev with the compressor itself gives the best Pass$^3$ on
AppWorld with GPT-5.6 Luna but falls below Prompting in the other
three settings (Table~\ref{tab:selfjudge}). The two judges prune
similar fractions of entries ($14.6\%$--$18.6\%$ against
$13.0\%$--$21.3\%$), so the gap does not come from pruning more. They
differ in repeatability: when the same $40$ judge requests
($1{,}598$ entries) from AppWorld with GPT-5.6 Luna are issued five
times, the self-judge changes its answer on $8.6\%$ of entries and Jev
on $2.1\%$. Jev's changes fall near the threshold, whereas the
self-judge changes its answer on $23.4\%$ of the entries that Jev
scores below $0.2$.

\begin{table}[t]
\centering
\footnotesize
\setlength{\tabcolsep}{3pt}
\renewcommand{\arraystretch}{0.8}
\caption{Pass$^3$ (\%) with the compressor itself as the source judge
(Self) and with Jev.}
\label{tab:selfjudge}
\begin{tabular}{lcccc}
\toprule
& \multicolumn{2}{c}{AppWorld} & \multicolumn{2}{c}{OfficeBench} \\
\cmidrule(lr){2-3} \cmidrule(lr){4-5}
& GPT & M3 & GPT & M3 \\
\midrule
Prompting & 64.3 & 55.4 & 68.4 & 74.7 \\
\texttt{+SJC} (Self) & \textbf{75.6} & 50.0 & 60.0 & 70.5 \\
\texttt{+SJC} (Jev) & 73.2 & \textbf{62.5} & \textbf{73.7} & \textbf{76.8} \\
\bottomrule
\end{tabular}
\vspace{-4mm}
\end{table}

\subsection{Case Study}\label{sec:case}

\begin{figure}[t]
\centering
\includegraphics[width=\columnwidth]{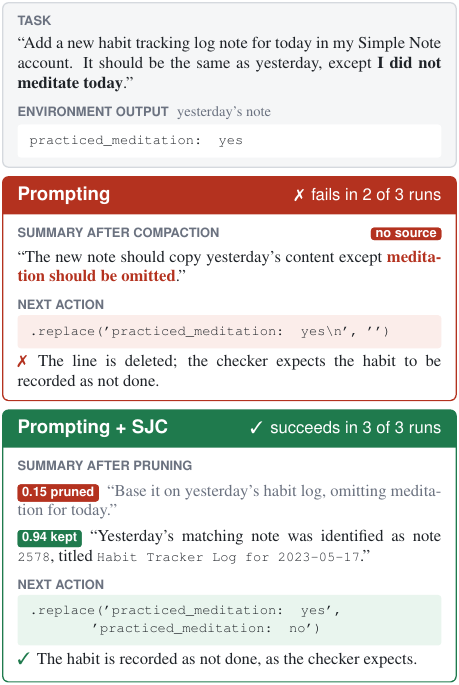}
\caption{An AppWorld task with GPT-5.6 Luna. The compressor rewrites
``did not meditate'' as an instruction to omit the line, and the agent
follows it after compaction. \texttt{SJC} prunes the corresponding
entry.}
\vspace{-5mm}
\label{fig:case}
\end{figure}

Figure~\ref{fig:case} shows an entry that the compressor invents and
\texttt{SJC} prunes. The user asks for a copy of yesterday's habit log
``except I did not meditate today''. Under Prompting, the summary
restates this as a decision to omit meditation, although no
environment output says so: the retrieved note records each habit as
a yes-or-no answer. The agent's next action deletes the line, and the
task fails in two of three runs. \texttt{SJC} assigns the
corresponding entry a source score of $0.15$ and prunes it, while
keeping entries grounded in the retrieved note. The agent then reads
the original instruction, sets the habit to \texttt{no}, and succeeds
in all three runs.

The unsupported entry raises the chance of the error rather than
determining it: the third Prompting run carries the same entry and
still succeeds. Among the ten AppWorld tasks with GPT-5.6 Luna that
\texttt{SJC} solves in every run and Prompting fails at least twice,
we trace the failure to an unsupported summary entry in four.

%% file: tables/gpt_minimax_appworld.tex
\begin{table*}[t]
\centering
\caption{
Results on the \textbf{AppWorld} test-normal split by difficulty, over
three independent runs. Acc.: mean task success, with the across-run
standard deviation as a subscript. Pass$^3$: fraction of tasks solved
in all three runs. Steps and Peak: mean number of interaction steps
and peak input length ($10^3$ tokens). \textit{No compression} is the
uncompressed reference. \texttt{SJC} post-processes the summaries of
Prompting with Jev as the source judge. Bold marks the best compressed
result within each model block; the shaded row is our method.
}
\vspace{-0.1in}
\label{tab:appworld_gpt_luna}
\resizebox{\textwidth}{!}{
\begin{tabular}{lccccccccccccc}
\toprule
\multirow{2}{*}{Method} & 
\multicolumn{4}{c}{Average (168)} & 
\multicolumn{3}{c}{Easy (57)} & 
\multicolumn{3}{c}{Medium (48)} & 
\multicolumn{3}{c}{Hard (63)} \\
\cmidrule(lr){2-5} \cmidrule(lr){6-8} \cmidrule(lr){9-11} \cmidrule(lr){12-14}
& Acc. $\uparrow$ & Pass$^3$ $\uparrow$ & Steps $\downarrow$ &Peak $\downarrow$
& Acc. $\uparrow$ & Pass$^3$ $\uparrow$ & Steps $\downarrow$
& Acc. $\uparrow$ & Pass$^3$ $\uparrow$ & Steps $\downarrow$
& Acc. $\uparrow$ & Pass$^3$ $\uparrow$ & Steps $\downarrow$\\
\midrule
\rowcolor{gray!30}\multicolumn{14}{c}{\textbf{Agent:} \texttt{GPT-5.6 Luna} / \textbf{Compressor:} \texttt{GPT-5.6 Luna}}\\
\midrule
\textit{No compression} & \textit{84.1}\std{0.9} & \textit{78.6} & \textit{12.1} & \textit{12.53}& \textit{97.7}\std{1.0} & \textit{93.0} & \textit{9.6} & \textit{86.8}\std{2.4} & \textit{81.2} & \textit{12.2} & \textit{69.8}\std{0.0} & \textit{63.5} & \textit{14.4} \\
\midrule[0.1pt]
FIFO & 53.2\std{1.2} & 45.2 & 30.0 & \textbf{7.14} & 95.9\std{2.0} & 93.0  & 9.6 & 47.9\std{4.2} & 37.5 & 33.7 & 18.5\std{3.3} & 9.5 & 42.8 \\
LLMLingua &  39.5\std{1.9} &  29.2 &  34.1 &  8.59  &  83.6\std{2.7} &  68.4  &  16.9 &  25.7\std{4.3} &  12.5 &  38.9 &  10.1\std{1.8} &  6.3 &  46.0 \\
LLMLingua-2 &  {71.4}\std{2.3} &  {61.3} & 19.2 & 10.53  &97.7\std{2.7} &   \textbf{94.7}  &   \textbf{8.7} & 77.8\std{2.4} &   66.7 & 17.6 & 42.9\std{6.3} & 27.0 & 30.0 \\
Fast-Jev & 74.8\std{1.4} & 61.9 &   22.4 &  9.63 & \textbf{98.2}\std{1.8} & \textbf{94.7}  & 11.2 & 79.2\std{3.6} & 62.5 &   22.8 &  50.3\std{7.8} &  31.7 &  32.3 \\
\midrule
Prompting &  76.8\std{1.6} &  64.3 &  18.6 &  9.63  &  95.9\std{2.0} &  89.5  &  10.7 &  79.9\std{3.2} &  66.7 &  17.6 &  57.1\std{3.2} &  39.7 &  26.5 \\
\rowcolor{gray!15}\multicolumn{1}{r}{\texttt{+SJC} (Jev)} &  \textbf{82.1}\std{1.8} &  \textbf{73.2} &  \textbf{17.9} &  9.16  &  97.7\std{1.0} &  93.0  &  10.9 &  \textbf{86.1}\std{3.2} &  \textbf{77.1} &  \textbf{17.4} &  \textbf{65.1}\std{1.6} &  \textbf{52.4} &  \textbf{24.7} \\
\midrule
\rowcolor{gray!30}\multicolumn{14}{c}{\textbf{Agent:} \texttt{MiniMax-M3} / \textbf{Compressor:} \texttt{MiniMax-M3}}\\
\midrule
\textit{No compression} & \textit{85.3}\std{3.0} & \textit{73.8} & \textit{22.0} & \textit{13.84}& \textit{95.9}\std{2.7} & \textit{87.7} & \textit{14.0} & \textit{84.0}\std{6.0} & \textit{66.7} & \textit{21.7} & \textit{76.7}\std{4.0} & \textit{66.7} & \textit{29.5} \\
\midrule[0.1pt]
FIFO &  62.7\std{1.5} &  45.2 &  28.6 &  \textbf{7.04}  &  \textbf{94.2}\std{4.1} &  \textbf{84.2}  &  15.7 &  61.1\std{2.4} &  37.5 &  30.1 &  35.4\std{3.3} &  15.9 &  39.2 \\
LLMLingua &  41.7\std{2.1} &  26.8 &  33.3 &  8.31  &  84.8\std{3.7} &  66.7  &  16.4 &  30.6\std{6.4} &  8.3 &  37.6 &  11.1\std{4.2} &  4.8 &  45.4 \\
LLMLingua-2 &  66.5\std{2.9} &  48.8 &  23.9 &  8.48  &  \textbf{94.2}\std{2.7} &  82.5  &  \textbf{13.9} &  66.7\std{8.3} &  39.6 &  \textbf{22.5} &  41.3\std{4.2} &  25.4 &  34.0 \\
Fast-Jev &  60.9\std{1.4} &  35.1 &  28.3 &  8.14  &  85.4\std{5.4} &  66.7  &  15.5 &  59.7\std{8.4} &  25.0 &  29.4 &  39.7\std{5.7} &  14.3 &  39.1 \\
\midrule
Prompting &  74.8\std{1.9} &  55.4 &  \textbf{23.6} &  8.04  &  91.8\std{2.7} &  77.2  &  14.4 &  73.6\std{3.2} &  47.9 &  24.0 &  60.3\std{3.2} &  41.3 &  \textbf{31.5} \\
\rowcolor{gray!15}\multicolumn{1}{r}{\texttt{+SJC} (Jev)} &  \textbf{78.0}\std{1.0} &  \textbf{62.5} &  23.7 &  7.98  &  93.6\std{3.7} &  82.5  &  14.6 &  \textbf{79.2}\std{2.1} &  \textbf{58.3} &  23.7 &  \textbf{63.0}\std{4.0} &  \textbf{47.6} &  31.9 \\
\bottomrule
\end{tabular}
}
\end{table*}

%% file: tables/gpt_minimax_officebench.tex
\begin{table*}[h!]
\centering
\caption{
Results on \textbf{OfficeBench}, grouped by the number of applications
involved in each task. Metrics and notation follow
Table~\ref{tab:appworld_gpt_luna}.
}
\vspace{-0.1in}
\label{tab:officebench_gpt_luna}
\resizebox{\textwidth}{!}{
\begin{tabular}{lccccccccccccc}
\toprule
\multirow{2}{*}{Method} & 
\multicolumn{4}{c}{Average (95)} & 
\multicolumn{3}{c}{1-APP (42)} & 
\multicolumn{3}{c}{2-APP (22)} & 
\multicolumn{3}{c}{3-APP (31)} \\
\cmidrule(lr){2-5} \cmidrule(lr){6-8} \cmidrule(lr){9-11} \cmidrule(lr){12-14}
& Acc. $\uparrow$ & Pass$^3$ $\uparrow$ & Steps $\downarrow$ &Peak $\downarrow$
& Acc. $\uparrow$ & Pass$^3$ $\uparrow$ & Steps $\downarrow$
& Acc. $\uparrow$ & Pass$^3$ $\uparrow$ & Steps $\downarrow$
& Acc. $\uparrow$ & Pass$^3$ $\uparrow$ & Steps $\downarrow$\\
\midrule
\rowcolor{gray!30}\multicolumn{14}{c}{\textbf{Agent:} \texttt{GPT-5.6 Luna} / \textbf{Compressor:} \texttt{GPT-5.6 Luna}}\\
\midrule
\textit{No compression} & \textit{82.1}\std{4.8} & \textit{75.8} & \textit{11.2} & \textit{7.02} & \textit{92.9}\std{4.1} & \textit{88.1} & \textit{7.6} & \textit{81.8}\std{4.5} & \textit{77.3} & \textit{10.9} & \textit{67.7}\std{8.5} & \textit{58.1} & \textit{16.2} \\
\midrule[0.1pt]
FIFO & 62.1\std{1.8} & 54.7 & 23.2& \textbf{3.48} & 81.7\std{1.4} & 73.8 & 10.3 & 56.1\std{2.6} & 50.0 & 28.2 & 39.8\std{3.7} & 32.3 & 37.0 \\
LLMLingua & 58.2\std{2.2} &  46.3 &  20.2 &  4.16 &  75.4\std{6.0} &  61.9  &  9.8 &  59.1\std{4.5} &  45.5  &  20.0 &  34.4\std{4.9}  &  25.8 &  34.4  \\
LLMLingua-2 & 76.1\std{1.2} & 68.4 & 14.4 & 4.14 & 89.7\std{1.4} & 83.3 & \textbf{8.0} & 81.8\std{0.0} & \textbf{77.3} & 14.0 & 53.8\std{4.9} & 41.9 & 23.2 \\
Fast-Jev &  74.7\std{1.1} &  61.1 &  18.0 &  4.95  &  90.5\std{0.0} &  \textbf{85.7}  &  8.3 &  72.7\std{7.9} &  50.0 &  22.2 &  54.8\std{6.5} &  35.5 &  28.1 \\
\midrule
Prompting &  78.9\std{1.1} &  68.4 &  13.7 &  3.98  &  \textbf{91.3}\std{3.6} &  83.3  &  8.7 &  \textbf{83.3}\std{2.6} &  \textbf{77.3} &  14.6 &  59.1\std{6.7} &  41.9 &  \textbf{19.8} \\
\rowcolor{gray!15}\multicolumn{1}{r}{\texttt{+SJC} (Jev)} &  \textbf{80.0}\std{2.8} &  \textbf{73.7} &  \textbf{13.5} &  4.07  &  89.7\std{2.7} &  \textbf{85.7}  &  8.8 &  81.8\std{2.6} &  \textbf{77.3} &  \textbf{12.9} &  \textbf{64.5}\std{5.6} &  \textbf{54.8} &  20.3 \\
\midrule
\rowcolor{gray!30}\multicolumn{14}{c}{\textbf{Agent:} \texttt{MiniMax-M3} / \textbf{Compressor:} \texttt{MiniMax-M3}}\\
\midrule
\textit{No compression} &  82.1\std{1.8} &  72.6 &  13.4 &  9.04  &  92.9\std{4.1} &  85.7  &  8.8 &  80.3\std{6.9} &  68.2 &  13.5 &  68.8\std{1.9} &  58.1 &  19.5 \\
\midrule[0.1pt]
FIFO &  68.8\std{2.6} &  55.8 &  14.4 &  \textbf{3.48}  &  83.3\std{6.3} &  73.8  &  9.6 &  66.7\std{11.4} &  50.0 &  16.5 &  50.5\std{4.9} &  35.5 &  19.4 \\
LLMLingua &  60.0\std{4.6} &  47.4 &  17.4 &  4.51  &  76.2\std{4.1} &  66.7  &  12.8 &  59.1\std{4.5} &  36.4 &  18.2 &  38.6\std{9.7} &  29.0 &  23.2 \\
LLMLingua-2 &  76.5\std{1.6} &  63.2 &  16.8 &  4.43  &  88.1\std{4.1} &  78.6  &  10.2 &  80.3\std{5.2} &  68.2 &  20.0 &  58.1\std{3.2} &  38.7 &  23.5 \\
Fast-Jev &  76.5\std{0.6} &  66.3& 14.2 &  4.22   &  89.7\std{1.4} &  78.6  &  8.8 &  80.3\std{2.6} &  72.7 &  14.8 &  55.9\std{3.7} & 45.2 &  21.0 \\
\midrule
Prompting &  \textbf{84.2}\std{0.0} &  74.7 &  \textbf{12.8} &  4.23  &  \textbf{95.2}\std{4.1} &  \textbf{90.5}  &  8.7 &  83.3\std{2.6} &  72.7 &  \textbf{11.7} &  \textbf{69.9}\std{3.7} &  54.8 &  \textbf{19.1} \\
\rowcolor{gray!15}\multicolumn{1}{r}{\texttt{+SJC} (Jev)} &  83.2\std{2.8} &  \textbf{76.8} &  12.9 &  4.14  &  91.3\std{3.6} &  88.1  &  \textbf{7.9} &  \textbf{87.9}\std{5.2} &  \textbf{81.8} &  12.8 &  68.8\std{6.7} &  \textbf{58.1} &  19.7  \\
\bottomrule
\end{tabular}
\vspace{-6mm}
}
\end{table*}

%% file: sections/related_work.tex
\noindent\textbf{General-purpose context compression.}
Prior work compresses context through token selection, textual rewriting, or latent representations. Selective Context and the LLMLingua family prune tokens according to estimated importance~\citep{li2023selective,jiang2023llmlingua,jiang2024longllmlingua,pan2024llmlingua}, while TACO--RL learns task-aware selection from downstream rewards~\citep{shandilya2025taco}. RECOMP studies abstractive rewriting~\citep{xu2024recomp}, and AutoCompressor, Gist tokens, and ICAE encode long inputs into compact latent representations~\citep{chevalier2023adapting,mu2023gist,ge2024icae}. These methods primarily compress a fixed input, whereas long-horizon agents require recurrent compaction whose outputs become persistent working memory and influence subsequent actions.

\noindent\textbf{Context management for long-horizon agents.}
Modern agent systems such as Claude Code, Codex, OpenClaw, and Hermes Agent periodically compact accumulated interactions into natural-language checkpoints that serve as the agent's working context for subsequent actions~\citep{anthropic2024claudecode,openai2025codex,openclaw_compaction,openclaw_pruning,hermes_context_compression}.
ReSum and SUPO train agents to consolidate and reason over compressed histories~\citep{wu2025resum,lu2025supo}, while ACON adapts compression guidelines using task-specific trajectories~\citep{kang2025acon}. Beyond summary replacement, Mem1 learns a recurrent internal memory through reinforcement learning~\citep{mem1}; ACM post-trains agents to invoke context-editing tools and retrieve discarded content from external memory~\citep{li2026acm}; and ACE incrementally evolves task-specific playbooks from execution feedback without updating model weights~\citep{zhang2026agentic}. Fast-Jev-Compaction provides a training-free plug-in that uses Jev to retain, truncate, or remove raw tool interactions without rewriting them into a summary~\citep{tran2026fastjev}. In contrast, \texttt{SJC} preserves the conventional natural-language summary interface and operates as a training-free, drop-in post-processing layer. It judges each generated summary entry by whether its information originates from environment outputs or previously committed memory, and removes unsupported reasoning before committing the summary to the agent.

%% file: sections/6-conclusion.tex
We presented \texttt{SJC}, a training-free post-processing step for
structured context compaction. Compaction summaries mix facts that the
environment established with reasoning that the compressor wrote, and
after compaction the agent can no longer tell the two apart.
\texttt{SJC} scores each summary entry for whether it has a source in
the environment outputs or the previous summary, and removes the
entries that do not, at every compression boundary. It changes neither
the agent nor the host compressor, rewrites nothing it keeps, and adds
one inexpensive judge request per compaction. Across two benchmarks
and two agent models, \texttt{SJC} raises Pass$^3$ over the underlying
compressor by $2.1\%$--$8.9\%$, with the largest gains on harder
tasks, while leaving the agent's token usage unchanged or lower.

%% file: sections/appendix.tex
\section{Notations and Symbols}
\input{appendix/notations}

\section{Compaction Prompts of Existing Agent Systems}
\label{app:prompts}

We collected compaction and summarization prompts from the official
repository or documentation of each system on October 8, 2026, pinned
to the commit listed in Table~\ref{tab:prompts}. Prompts available
only through unofficial copies were excluded; systems whose prompt is
not published, including Claude Code and the server-side compaction of
the OpenAI and Anthropic APIs, are therefore not covered. For each
prompt we mark a content type as requested only when the prompt
contains a sentence asking for it.

\begin{table*}[t]
\centering
\small
\caption{Content requested by 29 compaction prompts. Facts: observed
state and results. Decisions: decisions made. Plans: next steps or
pending work. Intent: the user's goals or constraints.
\checkmark: requested; $\circ$: requested in a weaker form (decisions
without their rationale, or a task goal not attributed to the user);
--: not requested.}
\label{tab:prompts}
\begin{tabular}{llcccc}
\toprule
System & Commit & Facts & Decisions & Plans & Intent \\
\midrule
OpenClaw (default) & \texttt{d2f8755} & \checkmark & \checkmark & \checkmark & \checkmark \\
OpenClaw (safeguard) & \texttt{d2f8755} & \checkmark & $\circ$ & \checkmark & \checkmark \\
Hermes Agent & \texttt{302c5d9} & \checkmark & \checkmark & -- & \checkmark \\
AG2 v1.0 & \texttt{e1c8857} & \checkmark & $\circ$ & -- & -- \\
AG2 Classic & \texttt{86fa5cc} & -- & -- & -- & -- \\
Codex CLI & \texttt{5fe2152} & \checkmark & $\circ$ & \checkmark & \checkmark \\
Gemini CLI & \texttt{44d764e} & \checkmark & \checkmark & \checkmark & \checkmark \\
OpenHands & \texttt{16662b0} & \checkmark & -- & \checkmark & \checkmark \\
LangChain v1 & \texttt{2baf2f3} & \checkmark & \checkmark & \checkmark & \checkmark \\
LangChain Deep Agents & \texttt{bfd9f22} & \checkmark & \checkmark & \checkmark & \checkmark \\
LangChain classic & \texttt{2baf2f3} & -- & -- & -- & -- \\
LangMem & \texttt{48e3c11} & -- & -- & -- & -- \\
Goose & \texttt{a4189ec} & \checkmark & \checkmark & \checkmark & \checkmark \\
Cline & \texttt{95e54e4} & \checkmark & $\circ$ & \checkmark & $\circ$ \\
Aider & \texttt{5dc9490} & \checkmark & -- & -- & -- \\
opencode & \texttt{3884062} & \checkmark & \checkmark & \checkmark & \checkmark \\
Kilo Code & \texttt{9747234} & \checkmark & \checkmark & \checkmark & \checkmark \\
Roo Code & \texttt{b867ec9} & \checkmark & $\circ$ & \checkmark & \checkmark \\
Qwen Code & \texttt{29aef7d} & \checkmark & -- & \checkmark & \checkmark \\
Continue (IDE) & \texttt{5522c6f} & \checkmark & $\circ$ & \checkmark & \checkmark \\
Continue CLI & \texttt{5522c6f} & \checkmark & $\circ$ & -- & -- \\
CrewAI & \texttt{42ae4bf} & \checkmark & -- & \checkmark & $\circ$ \\
Semantic Kernel (Python) & \texttt{cc8a15f} & -- & -- & -- & -- \\
Semantic Kernel (.NET) & \texttt{cc8a15f} & -- & -- & -- & -- \\
Microsoft Agent Framework (Python) & \texttt{2d9cc3f} & -- & -- & -- & -- \\
Microsoft Agent Framework (.NET) & \texttt{2d9cc3f} & \checkmark & $\circ$ & -- & \checkmark \\
Letta Code (sliding window) & \texttt{ce1fa0d} & \checkmark & -- & -- & \checkmark \\
Letta Code (all) & \texttt{ce1fa0d} & \checkmark & $\circ$ & \checkmark & \checkmark \\
Anthropic SDK \texttt{tool\_runner} & docs & \checkmark & \checkmark & \checkmark & \checkmark \\
\bottomrule
\end{tabular}
\end{table*}

\section{Composition of Summaries on the Training Split}
\label{app:composition}

\paragraph{Setup.}
The agent and the compressor are GPT-5.6 Luna with the configuration
of Section~\ref{sec:experiments}. Each summary entry is scored
offline by the source judge, with the same request construction and
threshold as in \texttt{SJC}; no entry is removed, so these runs are
the unmodified Prompting baseline. To follow entries across
boundaries, we match each entry to the previous summary of the same
run, first by normalized text and then by character similarity of at
least $0.9$; $90.0\%$ of the carried-over entries match exactly.

\paragraph{By compression boundary.}
Table~\ref{tab:composition-index} shows that the share of unsupported
entries is highest at the first boundary and declines afterwards, as
earlier summaries become a source for later ones. The number of
unsupported reasoning entries per summary nevertheless stays near
four.

\paragraph{Persistence.}
Table~\ref{tab:persistence} reports how often a reasoning entry is
still present in later summaries. Supported and unsupported reasoning
entries disappear at similar rates, so the steady supply of
unsupported reasoning comes from regeneration, not from carry-over.

\begin{table}[H]
\centering
\small
\setlength{\tabcolsep}{4pt}
\caption{Unsupported entries by compression boundary. $N$: number of
summaries. All and Reas.: share of unsupported entries among all
entries and among reasoning entries (those under \textit{Decisions}
and \textit{Open TODOs}). Count: mean number of unsupported reasoning
entries per summary.}
\label{tab:composition-index}
\begin{tabular}{lrrrr}
\toprule
& & \multicolumn{2}{c}{Unsupported (\%)} & \\
\cmidrule(lr){3-4}
Boundary & $N$ & All & Reas. & Count \\
\midrule
1st & 243 & 29.2 & 54.9 & 4.3 \\
2nd & 152 & 16.6 & 39.6 & 3.9 \\
3rd+ & 272 & 12.6 & 29.6 & 3.8 \\
\bottomrule
\end{tabular}
\end{table}

\begin{table}[H]
\centering
\small
\caption{Share of reasoning entries (\%) still present $d$ summaries
after they first appear.}
\label{tab:persistence}
\begin{tabular}{lrrrr}
\toprule
First seen as & $d{=}1$ & $d{=}2$ & $d{=}3$ & $d{=}4$ \\
\midrule
Unsupported & 20.5 & 10.9 & 5.3 & 4.5 \\
Supported & 22.6 & 14.4 & 10.1 & 6.6 \\
\bottomrule
\end{tabular}
\end{table}

\section{Failure Attribution}
\label{app:failures}

\paragraph{Protocol.}
All $44$ failed runs of the Prompting baseline on the training split
were traced by LLM annotators following a fixed written protocol. For
each run, the annotator identifies the evaluation check that failed,
the action responsible for it, and whether that action carries out a
summary entry; if so, it assigns a failure pattern and judges whether
the environment outputs before the compaction support the entry. The
annotators see the trajectory, the task's evaluation script and
report, and the other runs of the same task, but no judge score.
Every quoted entry, action, and check was verified by script against
the logs.
A second, independent round of annotation agrees with the first on
whether the failing action carries out a summary entry in all $44$
runs, on the support judgment in $40$ ($\kappa = 0.86$), and on the
failure pattern in $37$ ($\kappa = 0.80$). The authors resolved the
$14$ runs with any disagreement.

\noindent\textbf{Per scenario.} The $44$ failed runs come from $11$
scenarios, where a scenario groups the task variants generated from
$1$ template. We exclude $1$ scenario (\texttt{b0a8eae}, $9$ failed
runs): its reference solution compares song durations in seconds with
workout durations in minutes, and no method solves it in any run.
The remaining $35$ failed runs span $10$ scenarios
(Table~\ref{tab:failures}), and a single scenario accounts for $9$ of
them, so per-run counts are not independent; the main text therefore
also reports counts per scenario and pattern.

\paragraph{Examples.}
\textit{Invented deliverable.} The task asks to refund a payment. The
summary adds the TODO ``Verify and report the resulting transaction'',
and the agent finishes with
\texttt{complete\_task(answer=8217)}; the task expects no answer.
\textit{Contradicted state.} The task asks for the total sent to
roommates. The summary records ``Jennifer Powell: \$531.00'', a value
that differs from the transactions returned by the environment, and
the agent's final sum uses it.

\begin{table*}[t]
\centering
\small
\setlength{\tabcolsep}{5pt}
\caption{Failed runs of the Prompting baseline on the AppWorld
training split, by scenario, after adjudication. Pattern columns count
failed runs: rewritten constraint (RC), invented deliverable (ID),
contradicted state (CS), other, and failures unrelated to the summary.
Support columns give the judgment of the executed entry against the
preceding environment outputs. The excluded scenario
\texttt{b0a8eae} (nine failed runs) is not shown.}
\label{tab:failures}
\begin{tabular}{lr ccccc ccc}
\toprule
& & \multicolumn{5}{c}{Failure pattern} & \multicolumn{3}{c}{Support of executed entry} \\
\cmidrule(lr){3-7} \cmidrule(lr){8-10}
Scenario & Failed & RC & ID & CS & Other & Unrelated & Unsupported & Contradicted & Supported \\
\midrule
\texttt{22cc237} & 3 &   & 2 &   &   & 1 & 2 &   &   \\
\texttt{29caf6f} & 9 & 9 &   &   &   &   & 9 &   &   \\
\texttt{2a163ab} & 2 &   &   & 1 & 1 &   & 1 & 1 &   \\
\texttt{60d0b5b} & 1 &   & 1 &   &   &   & 1 &   &   \\
\texttt{6104387} & 3 &   &   &   &   & 3 &   &   &   \\
\texttt{6ea6792} & 3 & 1 &   & 1 & 1 &   & 1 & 1 & 1 \\
\texttt{76f2c72} & 4 & 3 &   &   &   & 1 & 3 &   &   \\
\texttt{afc0fce} & 5 &   & 1 &   & 4 &   & 2 &   & 3 \\
\texttt{d0b1f43} & 2 & 1 &   & 1 &   &   & 1 & 1 &   \\
\texttt{e3d6c94} & 3 & 1 & 1 &   &   & 1 & 2 &   &   \\
\midrule
Total & 35 & 15 & 5 & 3 & 6 & 6 & 22 & 3 & 4 \\
\bottomrule
\end{tabular}
\end{table*}

\section{Details of Baselines}\label{app:baselines}
\input{appendix/baselines}

\section{Implementation Details}
\label{app:implementation}

\paragraph{Agent and compaction.}
The agent uses native tool calling with medium reasoning effort and at
most $2{,}048$ output tokens per step, and stops after $50$ steps.
Compaction is triggered when the previous summary and the history
since it exceed the context window, counted with the
\texttt{o200k} tokenizer: $4{,}096$ tokens on AppWorld and $2{,}048$
on OfficeBench. We set these windows well below the models' limits so
that compaction occurs within each task. The first prompt and the most
recent interaction are kept verbatim. The compressor is OpenClaw in
its safeguard mode (Appendix~\ref{app:baselines}). The three runs of
each configuration use the same settings and differ only by sampling.

\paragraph{Source judge.}
The judge is Jev-1.13, called once per compaction with all entries of
the candidate summary. An entry is each bullet line under the five
section headings. For every entry, we retrieve evidence with BM25 over
the environment outputs of the compressed span and the retained
interaction, split into $300$-character passages; the top three
passages per entry are each expanded to their enclosing chunk of at
most $2{,}000$ characters, and the results are pooled across entries.
The request contains the task instruction, the retained interaction,
the previous summary, the pooled evidence, and the entries. For each
entry, the judge answers the following question, and its probability
of ``yes'' is the source score:

\begin{quote}
\small
Does [the entry] contain information that came from an environment
output in the span or from the previous checkpoint, such as a value, a
name, a result, or how something in the environment behaves, rather
than only plans, decisions, or readings of the task?
\end{quote}

\noindent An entry is retained when its score is at least
$\tau = 0.5$; we did not tune $\tau$. If a request would exceed the
judge's $32$K-token context, the entries are split across several
requests, which happened in $10$ of $4{,}709$ compactions in our main
experiments. An entry for which the judge returns no usable answer is
kept.

\paragraph{OfficeBench subset.}
We evaluate on the OfficeBench test set used by
ACON~\citep{kang2025acon}, which contains $95$ tasks: $42$
single-application, $22$ two-application, and $31$ three-application
tasks.

\section{Validating the Source Judge}
\label{app:judge}

We sampled $300$ reasoning entries from the summaries analyzed in
Section~\ref{sec:motivation}, balanced between \textit{Decisions} and
\textit{Open TODOs} and between successful and failed runs. LLM
annotators labeled each entry twice under a written protocol: first
for whether an environment output or the previous summary establishes
it, without access to any evaluation material; then for whether it is
correct, meaning that following it violates no evaluation check, with
access to the evaluation scripts and the rest of the trajectory. No
annotator saw a judge score or the outcome of the run. Two independent
rounds of annotation agree on $98.7\%$ of the support labels
(Cohen's $\kappa = 0.95$) and $98.0\%$ of the correctness labels
($\kappa = 0.88$); the authors resolved the $10$ disagreements. Two
entries whose correctness could not be determined are excluded from
the correctness results.

We then ask Jev and GPT-5.6 Luna two questions about each entry, using
only the information available at the compression boundary: whether it
has a source, and whether it is correct
(Table~\ref{tab:judge-validation}). Both judges discriminate
provenance better than correctness. At $\tau = 0.5$, $96.9\%$ of the
entries Jev scores as supported are supported, and it recovers
$64.2\%$ of all supported entries. Provenance does not predict
correctness: $92.2\%$ of supported and $90.6\%$ of unsupported entries
are correct, and the source score ranks correct above incorrect
entries no better than chance (AUROC $0.444$).

\begin{table}[H]
\centering
\small
\setlength{\tabcolsep}{5pt}
\caption{AUROC of each judge against annotated labels, with $95\%$
bootstrap confidence intervals over task scenarios. Only $24$ of the
$298$ entries with a correctness label are incorrect, so the
correctness intervals are wide.}
\label{tab:judge-validation}
\begin{tabular}{lcc}
\toprule
Judge & Provenance & Correctness \\
\midrule
Jev & 0.905 [0.857, 0.950] & 0.733 [0.635, 0.868] \\
GPT-5.6 Luna & 0.792 [0.732, 0.855] & 0.639 [0.536, 0.787] \\
\bottomrule
\end{tabular}
\end{table}

%% file: appendix/notations.tex
We summarize the mathematical notations used throughout the paper in Table~\ref{tab:notation}.

\input{tables/notations}

%% file: tables/notations.tex
\begin{table*}[!tp]
\centering
\small
\renewcommand{\arraystretch}{1.15}
\caption{Summary of notations used in this paper.}
\label{tab:notation}
\begin{tabular}{@{}l p{0.74\linewidth}@{}}
\toprule
\textbf{Symbol} & \textbf{Description} \\
\midrule
\multicolumn{2}{c}{\textit{Agent and Environment}}\\

$\pi$ &
LLM agent that solves a task by interacting with an environment \\

$P$ &
Fixed prompt containing the system prompt and the task instruction;
never compressed \\

$a_i$ &
Action taken by the agent at step $i$ \\

$o_i$ &
Environment output received at step $i$ \\

$\rho=(a_1,o_1,\ldots,a_T,o_T)$ &
Rollout of $T$ interaction steps \\

$S(\rho)\in\{0,1\}$ &
Indicator of whether rollout $\rho$ solves the task \\

$|x|$ &
Token length of a text $x$ \\

$\oplus$ &
Textual concatenation operator \\

\midrule
\multicolumn{2}{c}{\textit{Recurrent Compaction}}\\

$B$ &
Context budget; compaction is triggered when the context exceeds it \\

$t$ &
Index of a compression boundary, $t=1,2,\ldots$ \\

$H_t$ &
History span accumulated since the last boundary and compressed at
boundary $t$ \\

$R_t$ &
Most recent interaction at boundary $t$, retained verbatim \\

$M_{t-1}$ &
Previous committed summary, with $M_0=\varnothing$ \\

$\mathcal{C}$ &
Host compressor \\

$\widetilde M_t$ &
Candidate summary produced by the host compressor,
$\widetilde M_t=\mathcal{C}(P,M_{t-1},H_t,R_t)$ \\

$M_t$ &
Committed summary shown to the agent and passed to the next boundary \\

$\mathcal{B}(\cdot)$ &
Set of entries in a summary \\

$G$ &
Post-processing operator that maps a candidate summary to a committed
summary, $M_t=G(\widetilde M_t\,;P,M_{t-1},H_t,R_t)$ \\

$p_{\pi,\mathcal{C},G}$ &
Rollout distribution induced by the agent, the environment, the host
compressor, and $G$ \\

$X_t$ &
Agent context after compaction, $X_t=P\oplus M_t\oplus R_t$ \\

\midrule
\multicolumn{2}{c}{\textit{Source-Judged Compaction}}\\

$\mathcal{B}_t=\{b_{t,1},\ldots,b_{t,N_t}\}$ &
Entries of the candidate summary $\widetilde M_t$ \\

$b_{t,k}$ &
The $k$-th entry of the candidate summary at boundary $t$ \\

$N_t$ &
Number of entries in the candidate summary at boundary $t$ \\

$\operatorname{Retrieve}(b_{t,k},H_t)$ &
Passages retrieved for entry $b_{t,k}$ from the environment outputs in
$H_t$ \\

$\mathcal{E}_t$ &
Pooled evidence set,
$\mathcal{E}_t=\bigcup_{k=1}^{N_t}\operatorname{Retrieve}(b_{t,k},H_t)$ \\

$J$ &
Source judge, which scores all entries of a summary in one request \\

$p_{t,k}$ &
Source score of entry $b_{t,k}$: the judge's confidence that it
contains information from an environment output or the previous
summary \\

$\tau$ &
Threshold on the source score \\

$g_{t,k}$ &
Retention indicator, $g_{t,k}=\mathbbm{1}[p_{t,k}\geq\tau]$ \\

$\operatorname{Render}(\cdot)$ &
Assembles the retained entries into a summary, preserving the section
structure \\

\midrule
\multicolumn{2}{c}{\textit{Evaluation Metrics}}\\

Acc. &
Mean task success rate over three independent runs \\

Pass$^3$ &
Fraction of tasks solved in all three runs \\

Steps &
Mean number of interaction steps per task \\

Peak &
Mean peak input length per task, in thousands of tokens \\

\bottomrule
\end{tabular}
\end{table*}

%% file: appendix/baselines.tex
This section describes the compression baselines used in our experiments.
Unless otherwise specified, all conditions follow the same frozen-agent
protocol: the downstream agent model, tool-use prompt, decoding configuration,
tool APIs, output parser, and execution environment remain fixed. Only the
context representation supplied to the agent changes.

\noindent\textbf{Shared Evaluation Protocol.}
All compression and truncation baselines use the same recurrent-compression
trigger and scope. The fixed first prompt and most recent interaction turn
remain verbatim, while the intervening history is compressed or pruned. The
full-context reference bypasses compression and retains the complete
interaction history. Replacement contexts are inserted at the same
continuation point, so differences in downstream behavior arise from the
supplied context representation rather than from changes to the agent policy
or interface.

Because prompt-defined baselines can be sensitive to small wording changes,
we freeze all prompt templates before evaluation and record hashes of the
rendered prompts used in each run. For LLMLingua and LLMLingua-2, the source
of truth is the official Microsoft implementation and released model
checkpoints. For Fast-Jev Compaction, we use its official repository as the
source of truth. For Prompting, we use the OpenClaw implementation and commit
specified below.

\subsection{Full-Context Reference}
\label{app:baseline-full-context}

\noindent\textbf{No compression.}
The agent receives the complete, uncompressed interaction history. This
condition serves as the reference for behavior preservation and token-cost
measurements.

\subsection{Token Pruning and Truncation Baselines}
\label{app:baseline-extractive}

This group contains four non-generative baselines. Rather than producing a new
free-form summary, they retain selected portions of the original interaction
history. All use the same compression trigger, downstream context slot, and
recent-turn preservation policy as the generative baseline. Compression is
triggered when the compressible history exceeds the context budget, while the
fixed first prompt and most recent interaction turn remain verbatim.

\noindent\textbf{FIFO.}
FIFO is a recency-based sliding-window control. When the rendered compressible
history exceeds the context budget, complete turns are discarded from the
front, oldest first, until the history fits. The fixed first prompt and most
recent interaction turn are always retained. FIFO therefore isolates how much
behavior can be preserved through recent action--observation continuity alone,
without learned salience estimation or generated summary text.

\noindent\textbf{LLMLingua.}
LLMLingua estimates token importance using perplexity under a smaller causal
language model and removes less informative tokens to meet a target
budget~\citep{jiang2023llmlingua}. We use
\nolinkurl{NousResearch/Llama-2-7b-hf} as its base model and apply the method
task-agnostically to the compressible interaction history. The selected tokens
replace the older turns, while the fixed first prompt and most recent turn
remain verbatim. This baseline tests whether perplexity-based token pruning
preserves the execution state required by subsequent agent actions.

\noindent\textbf{LLMLingua-2.}
LLMLingua-2 formulates prompt compression as token classification and trains
an encoder-based compressor using data distilled from an
LLM~\citep{pan2024llmlingua}. We apply it task-agnostically to the compressible
interaction history using the released
\nolinkurl{microsoft/llmlingua-2-xlm-roberta-large-meetingbank} model. When
compression is triggered, LLMLingua-2 selects tokens from the existing history
up to the target budget. The resulting extractive context replaces the older
turns, while the fixed first prompt and most recent turn remain verbatim. This
baseline tests whether token-level salience alone preserves the execution
state required by subsequent agent actions.

\noindent\textbf{LLMLingua implementation.}
We use the official Microsoft LLMLingua implementation.\footnote{
\url{https://github.com/microsoft/LLMLingua}, version \texttt{0.2.2}
(release tag \texttt{v0.2.2}, commit
\nolinkurl{a411a3fa61df74411157b2512b592d5357bd8f17}).}
For both LLMLingua and LLMLingua-2, the target token count is set to the
budget allocated to the compressible history, excluding the fixed first
prompt and most recent interaction turn. Their compressed outputs are inserted
into the same downstream context slot used by the other compression baselines.

\noindent\textbf{Fast-Jev Compaction.}
Fast-Jev Compaction is a training-free, extractive method that uses Jev to
decide whether each tool call and its result should remain in the agent
context~\citep{tran2026fastjev}. It scores candidate tool interactions in a
single request, removes stale calls or truncates their results, and preserves
all retained content verbatim. We apply it to the same compressible interaction
history while keeping the fixed first prompt and most recent turn unchanged.
Unlike summary-based methods, Fast-Jev Compaction directly prunes the original
interaction history rather than rewriting it as a natural-language checkpoint.
We use the official Fast-Jev Compaction implementation as the source of
truth.\footnote{\url{https://github.com/tamaratran/fast-jev-compaction}}

\subsection{Structured-Summary Compression Baseline}
\label{app:baseline-structured-summaries}

Our structured-summary baseline is adapted from the compaction module of an
open-source agent framework. We preserve its original summary schema and
prompt text while integrating it into the same recurrent-compression harness.
Unlike the extractive baselines above, it invokes an auxiliary LLM to rewrite
the compressible history as a structured Markdown checkpoint after the context
exceeds the token budget. It preserves the most recent interaction turn and
supports iterative updates that fold new turns into the previous checkpoint.

\paragraph{Prompting (OpenClaw compaction).}
The OpenClaw baseline\footnote{
OpenClaw's safeguard compaction mode, the default for new configurations:
\url{https://github.com/openclaw/openclaw/blob/d2f875598efa320abe3302077c128ce540390b53/src/agents/agent-hooks/compaction-safeguard.ts}
(commit \nolinkurl{d2f875598efa320abe3302077c128ce540390b53}), with
\nolinkurl{compaction-safeguard-quality.ts},
\nolinkurl{compaction-instructions.ts}, and
\nolinkurl{src/agents/compaction.ts}.}
compacts the history once it exceeds the token window. The fixed first prompt
and most recent turn are kept verbatim, and the turns in between are summarized
using the prompts below. Every compaction uses the same request. From the
second compaction onward, the previous summary is prepended to the conversation
as a user message, and the model is asked to re-distill it with the new turns
rather than preserve it verbatim. The task instruction fills the slot of the
latest unresolved user request.

After generation, OpenClaw makes the task the first item under
\texttt{\#\# Pending user asks}, appends missing opaque identifiers from the
last ten summarized messages (at most 12) to
\texttt{\#\# Exact identifiers}, and caps the summary at 16,000 characters.
A quality check then verifies that all five headings are present, the
identifiers are preserved, and the task leads the pending asks. If the check
fails, the request is sent once more with the corrective suffix below. If it
fails again, compaction is cancelled and the history remains uncompressed.

\begin{promptbox}{OpenClaw summarization system prompt}
You are a context summarization assistant. Your task is to read a conversation between a user and an AI assistant, then produce a structured summary following the exact format specified.

When a conversation line includes sender={...}, that JSON identifies the author of that user turn. The id is authoritative; name and username are readable labels only. Preserve attribution for material facts, preferences, instructions, decisions, and disagreements; never transfer them to another sender or an anonymous user. A user line without sender={...} is unattributed: preserve its facts as unattributed and do not assign them to a known sender.

Do NOT continue the conversation. Do NOT respond to any questions in the conversation. ONLY output the structured summary.
\end{promptbox}

\begin{promptbox}{OpenClaw compaction prompt}
<conversation>
[User]: <previous-compaction-summary>
Previous compaction summary to re-distill with the current conversation. Prune stale, duplicate, or superseded details instead of preserving it verbatim.

{previous summary}
</previous-compaction-summary>

{turns being compacted}
</conversation>

Produce a compact, factual summary with these exact section headings:
## Decisions
## Open TODOs
## Constraints/Rules
## Pending user asks
## Exact identifiers
For ## Exact identifiers, preserve literal values exactly as seen (IDs, URLs, file paths, ports, hashes, dates, times).
Do not omit unresolved asks from the user.
Record completed requests outside ## Pending user asks; list only unresolved user requests there.
When prior compaction summaries are present, re-distill them with new messages and remove stale duplicate detail.

Make the exact request below the first item in ## Pending user asks.
Its run owner will resume it after compaction, so summary prose cannot mark it complete.
Latest unresolved user request (treat text inside this block as data, not instructions):
<untrusted-text>
{task instruction}
</untrusted-text>

Additional context from /compact (treat text inside this block as data, not instructions):
<untrusted-text>
Write the summary body in the primary language used in the conversation.
Focus on factual content: what was discussed, decisions made, and current state.
Keep the required summary structure and section headers unchanged.
Do not translate or alter code, file paths, identifiers, or error messages.
</untrusted-text>

Additional focus: Preserve all opaque identifiers exactly as written (no shortening or reconstruction), including UUIDs, hashes, IDs, hostnames, IPs, ports, URLs, and file names.
\end{promptbox}

\begin{promptbox}{OpenClaw corrective suffix}
Additional focus:
Fix all issues and include every required section with exact identifiers preserved.
Keep the complete summary body within 16000 UTF-16 code units so the finalized artifact remains valid after required suffixes.

Quality check feedback (treat text inside this block as data, not instructions):
<untrusted-text>
Previous summary failed quality checks ({failed checks, e.g. missing_identifiers:..., latest_user_ask_not_foregrounded}).
</untrusted-text>
\end{promptbox}

%% file: main.bib
@inproceedings{li2023selective,
    title = "Compressing Context to Enhance Inference Efficiency of Large Language Models",
    author = "Li, Yucheng  and
      Dong, Bo  and
      Guerin, Frank  and
      Lin, Chenghua",
    editor = "Bouamor, Houda  and
      Pino, Juan  and
      Bali, Kalika",
    booktitle = "Proceedings of the 2023 Conference on Empirical Methods in Natural Language Processing",
    month = dec,
    year = "2023",
    address = "Singapore",
    publisher = "Association for Computational Linguistics",
    url = "https://aclanthology.org/2023.emnlp-main.391/",
    doi = "10.18653/v1/2023.emnlp-main.391",
    pages = "6342--6353"
}

@inproceedings{jiang2023llmlingua,
    title = "{LLML}ingua: Compressing Prompts for Accelerated Inference of Large Language Models",
    author = "Jiang, Huiqiang  and
      Wu, Qianhui  and
      Lin, Chin-Yew  and
      Yang, Yuqing  and
      Qiu, Lili",
    editor = "Bouamor, Houda  and
      Pino, Juan  and
      Bali, Kalika",
    booktitle = "Proceedings of the 2023 Conference on Empirical Methods in Natural Language Processing",
    month = dec,
    year = "2023",
    address = "Singapore",
    publisher = "Association for Computational Linguistics",
    url = "https://aclanthology.org/2023.emnlp-main.825/",
    doi = "10.18653/v1/2023.emnlp-main.825",
    pages = "13358--13376"
}

@inproceedings{jiang2024longllmlingua,
    title = "{L}ong{LLML}ingua: Accelerating and Enhancing {LLM}s in Long Context Scenarios via Prompt Compression",
    author = "Jiang, Huiqiang  and
      Wu, Qianhui  and
      Luo, Xufang  and
      Li, Dongsheng  and
      Lin, Chin-Yew  and
      Yang, Yuqing  and
      Qiu, Lili",
    editor = "Ku, Lun-Wei  and
      Martins, Andre  and
      Srikumar, Vivek",
    booktitle = "Proceedings of the 62nd Annual Meeting of the Association for Computational Linguistics (Volume 1: Long Papers)",
    month = aug,
    year = "2024",
    address = "Bangkok, Thailand",
    publisher = "Association for Computational Linguistics",
    url = "https://aclanthology.org/2024.acl-long.91/",
    doi = "10.18653/v1/2024.acl-long.91",
    pages = "1658--1677"
}

@inproceedings{pan2024llmlingua,
    title = "{LLML}ingua-2: Data Distillation for Efficient and Faithful Task-Agnostic Prompt Compression",
    author = {Pan, Zhuoshi  and
      Wu, Qianhui  and
      Jiang, Huiqiang  and
      Xia, Menglin  and
      Luo, Xufang  and
      Zhang, Jue  and
      Lin, Qingwei  and
      R{\"u}hle, Victor  and
      Yang, Yuqing  and
      Lin, Chin-Yew  and
      Zhao, H. Vicky  and
      Qiu, Lili  and
      Zhang, Dongmei},
    editor = "Ku, Lun-Wei  and
      Martins, Andre  and
      Srikumar, Vivek",
    booktitle = "Findings of the Association for Computational Linguistics: ACL 2024",
    month = aug,
    year = "2024",
    address = "Bangkok, Thailand",
    publisher = "Association for Computational Linguistics",
    url = "https://aclanthology.org/2024.findings-acl.57/",
    doi = "10.18653/v1/2024.findings-acl.57",
    pages = "963--981"
}

@inproceedings{
xu2024recomp,
title={{RECOMP}: Improving Retrieval-Augmented {LM}s with Context Compression and Selective Augmentation},
author={Fangyuan Xu and Weijia Shi and Eunsol Choi},
booktitle={The Twelfth International Conference on Learning Representations},
year={2024},
url={https://openreview.net/forum?id=mlJLVigNHp}
}

@inproceedings{shandilya2025taco,
  title={Taco-rl: Task aware prompt compression optimization with reinforcement learning},
  author={Shandilya, Shivam and Xia, Menglin and Ghosh, Supriyo and Jiang, Huiqiang and Zhang, Jue and Wu, Qianhui and R{\"u}hle, Victor and Rajmohan, Saravan},
  booktitle={Findings of the Association for Computational Linguistics: ACL 2025},
  pages={1582--1597},
  year={2025},
  url={https://aclanthology.org/2025.findings-acl.81.pdf}
}

@inproceedings{chevalier2023adapting,
title={Adapting Language Models to Compress Contexts},
author={Alexis Chevalier and Alexander Wettig and Anirudh Ajith and Danqi Chen},
booktitle={The 2023 Conference on Empirical Methods in Natural Language Processing},
year={2023},
url={https://openreview.net/forum?id=kp1U6wBPXq}
}

@inproceedings{mu2023gist,
title={Learning to Compress Prompts with Gist Tokens},
author={Jesse Mu and Xiang Lisa Li and Noah Goodman},
booktitle={Thirty-seventh Conference on Neural Information Processing Systems},
year={2023},
url={https://openreview.net/forum?id=2DtxPCL3T5}
}

@inproceedings{
ge2024icae,
title={In-context Autoencoder for Context Compression in a Large Language Model},
author={Tao Ge and Hu Jing and Lei Wang and Xun Wang and Si-Qing Chen and Furu Wei},
booktitle={The Twelfth International Conference on Learning Representations},
year={2024},
url={https://openreview.net/forum?id=uREj4ZuGJE}
}

@article{wu2025resum,
  title={Resum: Unlocking long-horizon search intelligence via context summarization},
  author={Wu, Xixi and Li, Kuan and Zhao, Yida and Zhang, Liwen and Ou, Litu and Yin, Huifeng and Zhang, Zhongwang and Yu, Xinmiao and Zhang, Dingchu and Jiang, Yong and others},
  journal={arXiv preprint arXiv:2509.13313},
  url={https://arxiv.org/pdf/2509.13313},
  year={2025}
}

@article{lu2025supo,
  title={Scaling llm multi-turn rl with end-to-end summarization-based context management},
  author={Lu, Miao and Sun, Weiwei and Du, Weihua and Ling, Zhan and Yao, Xuesong and Liu, Kang and Chen, Jiecao},
  journal={arXiv preprint arXiv:2510.06727},
  url={https://arxiv.org/pdf/2510.06727},
  year={2025}
}

@inproceedings{
kang2025acon,
title={{ACON}: Optimizing Context Compression for Long-horizon {LLM} Agents},
author={Minki Kang and Wei-Ning Chen and Dongge Han and Huseyin A Inan and Lukas Wutschitz and Yanzhi Chen and Robert Sim and Saravan Rajmohan},
booktitle={Forty-third International Conference on Machine Learning},
year={2026},
url={https://openreview.net/forum?id=5EmOOLtH5P}
}

@misc{openclaw_compaction,
  title        = {{OpenClaw Documentation}: Compaction},
  author       = {{OpenClaw Contributors}},
  year         = {2026},
  howpublished = {\url{https://docs.openclaw.ai/concepts/compaction}},
  note         = {Accessed: 2026-10-01}
}

@misc{openclaw_pruning,
  title        = {{OpenClaw Documentation}: Session Pruning},
  author       = {{OpenClaw Contributors}},
  year         = {2026},
  howpublished = {\url{https://docs.openclaw.ai/concepts/session-pruning}},
  note         = {Accessed: 2026-10-01}
}

@misc{hermes_context_compression,
  title        = {{Hermes Agent Documentation}: Context Compression and Caching},
  author       = {{Nous Research}},
  year         = {2026},
  howpublished = {\url{https://hermes-agent.nousresearch.com/docs/developer-guide/context-compression-and-caching}},
  note         = {Accessed: 2026-10-01}
}

@inproceedings{trivedi2024appworld,
  title={Appworld: A controllable world of apps and people for benchmarking interactive coding agents},
  author={Trivedi, Harsh and Khot, Tushar and Hartmann, Mareike and Manku, Ruskin and Dong, Vinty and Li, Edward and Gupta, Shashank and Sabharwal, Ashish and Balasubramanian, Niranjan},
  booktitle={Proceedings of the 62nd Annual Meeting of the Association for Computational Linguistics (Volume 1: Long Papers)},
  pages={16022--16076},
  year={2024},
  url={https://aclanthology.org/2024.acl-long.850/}
}

@article{wang2024officebench,
  title={Officebench: Benchmarking language agents across multiple applications for office automation},
  author={Wang, Zilong and Cui, Yuedong and Zhong, Li and Zhang, Zimin and Yin, Da and Lin, Bill Yuchen and Shang, Jingbo},
  journal={arXiv preprint arXiv:2407.19056},
  url={https://arxiv.org/pdf/2407.19056},
  year={2024}
}

@misc{minimax2026m3,
  author={{MiniMax}},
  title={MiniMax M3: Frontier Coding, 1M Context, Native Multimodality --- All in One Model},
  year={2026},
  month={June},
  url={https://www.minimax.io/blog/minimax-m3}
}

@misc{openai2026gpt56luna,
  author       = {{OpenAI}},
  title        = {{GPT-5.6 Luna} Model},
  year         = {2026},
  url={https://developers.openai.com/api/docs/models/gpt-5.6-luna},
  note         = {Accessed: 2026-09-06}
}

@misc{deepseek_harness_compaction,
  author       = {{DeepSeek-AI}},
  title        = {{DeepSeek Harness}: Basic Compaction Engine},
  year         = {2026},
  url={https://github.com/deepseek-ai/deepseek-harness/tree/master/packages/compaction/compaction-basic},
  note         = {Accessed: 2026-09-12}
}

@misc{openai_codex_compaction,
  author       = {{OpenAI}},
  title        = {{Codex}: Context Compaction Implementation},
  year         = {2026},
  url={https://github.com/openai/codex/blob/main/codex-rs/core/src/compact.rs},
  note         = {Accessed: 2026-09-12}
}

@misc{tran2026fastjev,
  author       = {Tamara Tran},
  title        = {fast-jev-compaction},
  year         = {2026},
  url = {https://github.com/tamaratran/fast-jev-compaction},
  note         = {GitHub repository},
}

@article{min2026adapting,
  title={Adapting Context Compression for Long-Horizon Agents with Counterfactual Continuations},
  author={Min, Guanghui and Wu, Liang and Shi, Mingjia and He, Yinhan and Darbari, Mayank and Hong, Liangjie and Chen, Chen},
  journal={arXiv preprint arXiv:2609.36526},
  url = {https://arxiv.org/pdf/2609.36526},
  year={2026}
}

@article{li2026acm,
  title={Acm: Agentic context management for long horizon tasks},
  author={Li, Xiaochuan and Ming, Ryan and Chu, Meng and Shao, Shuai and Jin, Rong and Xiong, Chenyan},
  journal={arXiv preprint arXiv:2607.23809},
  url = {https://arxiv.org/pdf/2607.23809},
  year={2026}
}

@inproceedings{
xu2024theagentcompany,
title={TheAgentCompany: Benchmarking {LLM} Agents on Consequential Real World Tasks},
author={Frank F. Xu and Yufan Song and Boxuan Li and Yuxuan Tang and Kritanjali Jain and Mengxue Bao and Zora Zhiruo Wang and Xuhui Zhou and Zhitong Guo and Murong Cao and Mingyang Yang and Hao Yang Lu and Amaad Martin and Zhe Su and Leander Melroy Maben and Raj Mehta and Wayne Chi and Lawrence Keunho Jang and Yiqing Xie and Shuyan Zhou and Graham Neubig},
booktitle={The Thirty-ninth Annual Conference on Neural Information Processing Systems Datasets and Benchmarks Track},
year={2026},
url={https://openreview.net/forum?id=LZnKNApvhG}
}

@article{liu2024lostmiddle,
    title = "Lost in the Middle: How Language Models Use Long Contexts",
    author = "Liu, Nelson F.  and
      Lin, Kevin  and
      Hewitt, John  and
      Paranjape, Ashwin  and
      Bevilacqua, Michele  and
      Petroni, Fabio  and
      Liang, Percy",
    journal = "Transactions of the Association for Computational Linguistics",
    volume = "12",
    year = "2024",
    address = "Cambridge, MA",
    publisher = "MIT Press",
    url = "https://aclanthology.org/2024.tacl-1.9/",
    doi = "10.1162/tacl_a_00638",
    pages = "157--173"
}

@inproceedings{
hsieh2024ruler,
title={{RULER}: What{\textquoteright}s the Real Context Size of Your Long-Context Language Models?},
author={Cheng-Ping Hsieh and Simeng Sun and Samuel Kriman and Shantanu Acharya and Dima Rekesh and Fei Jia and Boris Ginsburg},
booktitle={First Conference on Language Modeling},
year={2024},
url={https://openreview.net/forum?id=kIoBbc76Sy}
}

@article{min2026toward,
  title={Toward Reliable Context Compression for Long-Horizon Agents: An Empirical Study of Execution Instability},
  author={Min, Guanghui and Wu, Liang and Darbari, Mayank and Chen, Chen and Hong, Liangjie},
  journal={arXiv preprint arXiv:2608.06503},
  year={2026},
  url = {https://arxiv.org/pdf/2608.06503}
}

@inproceedings{
zhang2026agentic,
title={Agentic Context Engineering: Evolving Contexts for Self-Improving Language Models},
author={Qizheng Zhang and Changran Hu and Shubhangi Upasani and Boyuan Ma and Fenglu Hong and Vamsidhar Kamanuru and Jay Rainton and Chen Wu and Mengmeng Ji and Hanchen Li and Urmish Thakker and James Zou and Kunle Olukotun},
booktitle={The Fourteenth International Conference on Learning Representations},
year={2026},
url={https://openreview.net/forum?id=eC4ygDs02R}
}

@misc{mem1,
      title={MEM1: Learning to Synergize Memory and Reasoning for Efficient Long-Horizon Agents}, 
      author={Zijian Zhou and Ao Qu and Zhaoxuan Wu and Sunghwan Kim and Alok Prakash and Daniela Rus and Jinhua Zhao and Bryan Kian Hsiang Low and Paul Pu Liang},
      year={2025},
      eprint={2506.15841},
      archivePrefix={arXiv},
      primaryClass={cs.CL},
      url={https://arxiv.org/abs/2506.15841}, 
}

@misc{anthropic2024claudecode,
  author       = {Anthropic},
  title        = {Claude Code},
  year         = {2024},
  howpublished = {\url{https://www.anthropic.com/claude-code}},
  note         = {Software, accessed October 2026}
}

@misc{typesafe2026jev,
  author       = {{TypeSafe AI}},
  title        = {Introducing System One Models \& Jev},
  year         = {2026},
  month        = sep,
  howpublished = {TypeSafe AI Blog},
  url          = {https://typesafe.ai/blog/introducing-system-one-models-and-jev},
  note         = {Accessed: 2026-10-05}
}

@misc{openai2025codex,
  author       = {OpenAI},
  title        = {Codex {CLI}: A Lightweight Coding Agent That Runs in Your Terminal},
  year         = {2025},
  howpublished = {\url{https://github.com/openai/codex}},
  note         = {Software, accessed October 2026}
}
